\documentclass[11pt,a4paper]{article}
\usepackage[hyperref]{acl2020}
\usepackage{times}
\usepackage{latexsym}

\usepackage{graphicx}
\usepackage{booktabs}

\usepackage{microtype}
\usepackage{amsfonts}
\aclfinalcopy 

\title{Muppet: Massive Multi-task Representations with Pre-Finetuning}
\author{Armen Aghajanyan \\
  Facebook \\
  \texttt{armenag@fb.com} \\\And
  Anchit Gupta \\
  Facebook \\
  \texttt{anchit@fb.com} \\\And
  Akshat Shrivastava \\
  Facebook \\
  \texttt{akshats@fb.com} \\ \And
  Xilun Chen \\
  Facebook \\
  \texttt{xilun@fb.com} \\ \AND
  Luke Zettlemoyer \\
  Facebook \\
  \texttt{lsz@fb.com} \\\And
  Sonal Gupta \\
  Facebook \\
  \texttt{sonalgupta@fb.com} \\}

\date{}

\begin{document}
\maketitle
\begin{abstract}
We propose pre-finetuning, an additional large-scale learning stage between language model pre-training and fine-tuning. Pre-finetuning is massively multi-task learning  (around 50 datasets, over 4.8 million total labeled examples), and is designed to encourage learning of representations that generalize better to many different tasks.
We show that pre-finetuning 
consistently improves performance for pretrained discriminators (e.g.~RoBERTa) and generation models (e.g.~BART)  on a wide range of tasks (sentence prediction, commonsense reasoning, MRC, etc.), while also significantly improving sample efficiency during fine-tuning. 
We also show that large-scale multi-tasking is crucial; pre-finetuning can hurt performance when few tasks are used up until a critical point (usually above 15) after which performance improves linearly in the number of tasks. 
\end{abstract}
\section{Introduction}



The recent success of language model pre-training \cite{BERT, ROBERTA, BART, T5, GPT} is remarkable, at least in part, due to the exclusive use of self supervision, without any manually labeled data. For many tasks, however, we already have training examples for related problems, which we should be able to leverage. Recent work has shown gains from fine-tuning schemes that are multi-task~\cite{T5, unifiedqa} and multi-stage~\cite{MT_DNN}, but it can be difficult to know which intermediate tasks will best transfer~\cite{T5}. In this paper, we show that multi-task supervised tuning, if done at a sufficiently large scale with many different tasks, can be an effective second stage of task-agnostic pre-training, removing the need to pre-select the best intermediate tasks.

More specifically, 
in addition to the standard pre-training/fine-tuning methodology of learning language tasks, we introduce a new intermediate stage, \textit{pre-finetuning}.
Pre-finetuning involves a massive multi-task learning step (4.8 million total training examples) performed on around 50 classification, summarization, question answering, and common sense reasoning tasks. We believe we are the first to investigate multi-task learning at this scale in terms of both number and types of tasks. We show, in particular, that standard multi-tasking schemes can be unstable and often fail to learn high quality representations.  However, we introduce a new training scheme which uses loss scaling and task-heterogeneous batches so that gradient steps are more evenly balanced across multiple different competing tasks, greatly improving training stability and overall performance. We call our pre-finetuned models MUPPET; Massive Multi-task RePresentation with PrE-fineTuning.

Through extensive experiments, we show that incorporating pre-finetuning to RoBERTa \cite{ROBERTA} and BART \cite{BART} models yields consistent improvements, including new state-of-the-art performance for RTE \cite{rte} and HellaSWAG \cite{hellaswag}, without having to specify specific intermediate transfer tasks. These gains are particularly strong in the low resource regime, where there is relatively little labeled data for fine-tuning.  
We also study why pre-finetuning outperforms previous multi-tasking schemes. We first compare different optimization techniques to stabilize training, and find it important to use task-heterogeneous batches with task-rebalancing loss scaling. We also show that scale is crucial for effective multi-task learning.
We empirically see a critical point in terms of the number of tasks (usually over 15); having fewer tasks degrades representations, while having more seems to improve performance linearly as far as we were able to scale.

To summarize, our contributions include:
\begin{itemize}
    \item We show that we can further improve pre-trained representations with an additional stage we call pre-finetuning, which utilizes massively multi-task learning. We show standard pre-trained representations, when further refined with pre-finetuning consistently improve performance on downstream tasks.
    \item We introduce a new multi-task training scheme for effective learning at scale, which uses loss scaling and task-heterogeneous batches.
    \item We explore the effects of scale on multi-task learning and show the existence of critical points in multi-task training, beyond which increasing the number of tasks improves generalizable representations.
    \item We conduct a study surrounding the data efficiency of standard pre-trained representations and their respective pre-finetuned counterparts. We show that the pre-finetuned models consistently require less data for fine-tuning.
\end{itemize}
\section{Related Work}
Multi-task learning has been an increasingly active topic in recent literature. Recent advances such as MT-DNN show that by leveraging multi-task learning, we can further improve performance on several language benchmarks on top of traditional pre-training \citep{MT_DNN}. However, T5 \citep{T5} shows that incorporating multi-task learning ontop of larger models does not improve upon the standardized pre-training / finetuning. Thus the effect of multi-task learning across different pre-training methods is not fully understood. 

Recently \citet{unifiedqa} showed how doing MTL training on a range of QA tasks can improve the performance of T5 by taking advantage of cross dataset transfer. Unlike our approach, they convert all the data to a seq2seq format, operate on a smaller MTL scale, have a different batching strategy, and focus solely on improving QA tasks. Our work shows how even seemingly very different datasets, for example, summarization and extractive QA, can help each other by improving the model's representations.

Our work aims to explore multi-task learning at a much larger scale; by incorporating a larger number of tasks, we show that we can consistently improve several language benchmarks from several domains. Contrary to T5, we show that incorporating a secondary stage of multi-task learning does lead to better representations. In \S \ref{sec:understanding_mtl} we demonstrate the effectiveness of multi-task learning to be coming from the large scale of our MTL setup.

\section{Pre-Finetuning Through Massive Multitask Learning}

Previous work has reported mixed results from experiments on multi-task learning~\cite{MT_DNN, T5}. In general, it can be challenging to balance the losses from different tasks; upsampling can lead to overfitting low resource tasks, and downsampling can lead to improper learning of specific tasks. This difficulty is particularly pronounced when operating at the scale of experiments we show in Section~\ref{sec:scale_ablation}, where there are more diverse tasks than previously considered. This section presents our pre-finetuning approach that leads to more stable and accurate multi-task training by introducing new optimization, loss scaling, and task sampling schemes to balance each minibatch's updates better.

\subsection{Tasks and Losses}

\paragraph{Diverse Tasks}
To learn general language representations, we include a variety of tasks across many domains. We select language tasks across four different domains: \textit{classification}, \textit{commonsense reasoning}, \textit{machine reading comprehension}, and \textit{summarization}. In Table~\ref{tab:mtl_datasets}, we show the break down of each of the task types along with the number of samples used from each during pre-finetuning. In total our multi-task set up learns over 4.8 supervised samples across 4 families of tasks.

\begin{table}[]
\centering
\small
\begin{tabular}{@{}lrrrr@{}}
\toprule
Task Type                     & \# Datasets & \# Train & \# Eval  \\  
\midrule
Classification                &    26         &    2.9M      &        188K         \\
Summarization                &   4          &      524K     &           30K      \\
MRC &   6           &     1.05M     &    123M            \\
CommonSense         &         10    &          360K & 49K\\   \midrule 
Total                &  46           &    4.8M    &  390K \\ 
\bottomrule
\end{tabular}
\caption{Break down of MTL pre-finetuning datasets. The table shows the number of datasets we used per task type and the number of samples in training and evaluation sets.}
\label{tab:mtl_datasets}
\end{table}

A full list of all of the datasets we leverage for pre-finetuning is described in appendix \S \ref{sec:appendix_dataset}.

\paragraph{Standard Losses} 
To train on several datasets, our model contains task-specific heads, each optimizing for a task-specific loss. The loss functions are summarized in table \ref{tab:mtl_loss}. Each loss is scaled with loss scaling described in \S \ref{sec:loss_scaling}. After loss scaling, the gradients from each task are averaged before doing the model update step.

\begin{table}[]
\centering
\small
\begin{tabular}{@{}ll@{}}
\toprule
Task Type                      & Loss Function  \\  \midrule
Classification                &    Cross Entropy (CE)              \\
Summarization                 &   Label Smoothed CE \cite{inception_label_smoothing}                \\
MRC &     Span Prediction \cite{bidaf}        \\
Commonsense         &  Sentence Ranking Loss  \cite{ROBERTA}         \\
\bottomrule
\end{tabular}
\caption{Description of loss functions for each task type. Note for summarization the label smoothed cross entropy loss is averaged across tokens.}
\label{tab:mtl_loss}
\end{table}

\subsection{Optimization}

We show two strategies to learn multi-task representations at scale: \textit{Accumulating Gradients Across Tasks (Heterogeneous Batches)} and \textit{Leveraging Better Finetuning}.

\paragraph{Accumulating Gradients Across Tasks} Our model is trying to optimize not a single objective but several potentially competing objectives to create a unified representation across several tasks during model training. During gradient descent, moving along the gradient of a single task may not be the optimal direction for the model to move to learn a single unified representation across tasks. To overcome this, we ensure each batch our model optimizes consists of several tasks. Each worker samples a random batch from our set of tasks and computes a gradient, accumulated for the final update. Empirically we use 64 GPUs for pre-finetuning, resulting in each batch consisting of gradients across 64 sampled tasks. In \S\ref{sec:batches} we show how such a strategy allows for our model to arrive at a better representation for end task finetuning.

\paragraph{Better Finetuning} 
Instead of starting from scratch, we initialize our model with representations learned from self-supervised pre-training in pre-finetuning. This can inherit the knowledge captured in the pre-trained representations and speed up training. \citet{stability_bert} show that standard fine-tuning of pre-trained models can be unstable, which may be aggravated in our case as we are training on a diverse set of tasks simultaneously. Therefore, we employ the R3F/R4F methods~\cite{RXF} to combat this issue. In particular, R3F/R4F consists of an additional loss term, ensuring that small perturbations to the input space result in similar representations, which can be used to learn more robust representations during pre-finetuning.

In early experimentation, we found that R3F was pivotal in getting MUPPET to work for BART. All other fine-tuning and pre-finetuning was done using standard SGD.

\subsection{Loss Scaling}
\label{sec:loss_scaling}
Loss scaling methods introduce a multiplicative reweighting of individual losses per data-point. Various loss scaling techniques have been proposed, from dynamic scaling by inverse training loss to simple scaling by the number of data-points in respective datasets \citep{gradnorm_mtl}.

As pre-finetuning optimizes several different types of tasks and datasets, each having its own output spaces, loss scaling becomes essential to ensure stable training.
We attempted various forms of loss-scaling throughout initial experimentation, but the most effective was the novel method we describe below. 

Let us denote $\mathcal{L}_i(x_i, y_i; \theta)$ as the loss for datapoint $i$ for a model parameterized by $\theta$. Remember that the loss depends on the type of task (commonsense loss is different from binary classification). Furthermore let $n: \mathbb{N} \rightarrow \mathbb{N}$ be a function which for each data-point returns the number of predictions $\mathcal{L}$ operates over. For example, for binary classification, $n$ would return two, while for generation, $n$ would return the size of the vocabulary (since we average across loss per token generated). We scale data-point loss so that, if the class distribution were uniformly distributed along with our models predictions, all of our losses would have equivalent values.
\begin{equation}
    \mathcal{L}^{scaled}_i(x_i, y_i; \theta) = \frac{\mathcal{L}_i(x_i, y_i; \theta)}{\log{n(i)}}
\end{equation}

We found that this static scaling worked incredibly well, outperforming other loss scaling methods in early experimentation.

\subsection{Sampling}
Another approach to balancing various tasks in a multi-task set up is to up-sample smaller datasets and down-sample larger ones to achieve more uniformity between dataset sizes.

Existing results for dataset sampling methods in multi-task learning are conflicting, but recent work has shown that it does not work well for multi-task learning of pre-trained representations. For example, T5 showed that all various forms of sampling did not improve overusing the natural size of datasets \citep{T5}.

We also found that sampling datasets were consistently detrimental for multi-task learning over pre-trained representations during initial experimentation. Specifically, we saw unmanageable over-fitting and stability issues. Therefore we opt for maintaining the natural distribution of the datasets throughout all of our experiments.

\subsection{Experimental Setup}
We selected RoBERTa \citep{ROBERTA} and BART \citep{BART} as our initial pre-trained models to further pre-finetune. For each task type we use a different prediction scheme.
Every \textit{Sentence Prediction} dataset gets a separate classification head, for \textit{Commonsense} and \textit{MRC} we utilize a separate unified head for each task. For \textit{Summarization}, we do not add any parameters and use the BART decoder and output layer as is. Experimentally we saw using a different head per individual \textit{Commonsense} and \textit{MRC} datasets lead to severe overfitting. 

For both models, we do the pre-finetuning procedure for both the \textit{Base} and \textit{Large} models. We trained each model configuration with 64 GPUs until convergence. Dependent on configuration, this ranged from a day to 4 days. We include the hyper-parameters used per pre-finetuning run in the Appendix in Section~\S\ref{sec:appendix_hp}.

\section{Empirical Results}
\label{sec:empirical_results}
We first show that pre-finetuning improves the representations of pre-training models. To do so, we fine-tune our pre-finetuned models on a large set of tasks. 

For each of the individual downstream tasks, we use a fixed hyper-parameter search to optimize over simple hyperparameters such as learning rate, Adam $\epsilon$ \citep{ADAM} and dropout \citep{DROPOUT}. We present our results in two tables. Table~\ref{table:mtl_glue_mrc} shows our results on the GLUE benchmark \citep{GLUE} as well as two \textit{MRC} tasks; SQuAD \citep{SQUAD} and ReCoRD \citep{RECORD}. Table~\ref{table:mtl_sp_commonsense} reports results on other \textit{Sentence Prediction} tasks as well as \textit{Commonsense} tasks. We also include results from MT-DNN \cite{MT_DNN}, ELECTRA \cite{ELECTRA},\footnote{For ELECTRA results we leverage the results presented in the ELECTRA github \url{https://github.com/google-research/electra\#expected-results}} and RoBERTa \cite{ROBERTA} models. For \textit{Summarization} tasks we show that our pre-finetuned BART model outperforms all other summarization baselines. Both of these tables report over data-sets available during the pre-finetuning stage.

Given that our pre-finetuned models now have an understanding of the task at hand through the use of classification heads, we have a choice during finetuning on whether or not to use these heads. In general we found re-using heads to be beneficial for \textit{MRC}, \textit{Commonsense} and \textit{Sentence Prediction} tasks with small dataset size.
\begin{table*}[th]
\centering
\small
\begin{tabular}{@{}llllllllll@{}}
\toprule
 & \multicolumn{6}{c}{GLUE} & \multicolumn{1}{c}{MRC} \\
 \cmidrule(lr){2-7}\cmidrule(lr){8-9} 
 & 
  MNLI &
  QQP &
  RTE &
  QNLI &
  MRPC &
  SST-2 &
  SQuAD &\\ \midrule
RoBERTa-B   & 87.6    & \textbf{91.9}    & 78.7 & 92.8 & 90.2   &94.8 & 82.6   \\
\quad\quad + MUPPET       & \textbf{88.1}         &  \textbf{91.9}         &  \textbf{87.8}   &     \textbf{93.3}  &   \textbf{91.7}      &  \textbf{96.7}    &  \textbf{86.6}    \\ \\ 
RoBERTa-L & 90.2   & \textbf{92.2}    & 88.1    &  94.7    & 90.9   &   96.4    & 88.7     \\
\quad\quad  + MUPPET        & \textbf{90.8}  &   \textbf{92.2}       &  \textbf{\underline{92.8}}    &   \textbf{94.9}   & \textbf{91.4}      &   \textbf{\underline{97.4}} &  \textbf{\underline{89.4}} \\ \\
BART   &  \textbf{89.9}    & 92.5   & 87.0 & \textbf{94.9} & 90.4 & 96.6 &    \\
\quad\quad  + MUPPET       &    \textbf{89.9}       &   \textbf{\underline{92.7}}       &  \textbf{92.4}  &  94.6     &   \textbf{\underline{92.2}}      & \textbf{96.9}     &     \\
\midrule
ELECTRA-B & 88.8 & 91.5 & 82.7 & 93.2 & 89.5 & 95 & 80.5 \\  
ELECTRA-L  & \underline{90.9} & 92.4 & 88.0 & \underline{95.0} & 90.8 & 96.9 & 88.1 \\ 
MT-DNN & 87.1 & 91.9/89.2 & 83.4 & 92.9 & 91.0/87.5 & 94.3 &- \\ \bottomrule
\end{tabular}
\caption{We present results for the GLUE benchmark task and a MRC dataset. Bolded numbers show the MUPPET vs. base model, underline marks the best number. If not explicitly stated, the results are showing the accuracy of the evaluation set. For the MRC tasks, we report both exact match (EM) and F1 as is standard in the literature. For SQuAD, we reused the task head from pre-finetuning.}
\label{table:mtl_glue_mrc}

\end{table*}

\begin{table*}[th]
\centering
\small
\begin{tabular}{@{}lllll|lllll@{}}
\toprule
 & \multicolumn{1}{c}{SP} & \multicolumn{3}{c}{Commonsense} & \multicolumn{3}{c}{Summarization}\\
 \cmidrule(lr){2-2}\cmidrule(lr){3-5}\cmidrule(lr){6-8} 
 & 
  BoolQ &
  CQA &
  HellaSwag &
  OpenQA & 
  CNN/DailyMail & 
  Gigaword      & 
  Reddit TIFU\\ \midrule

RoBERTa-B   & 82.0       & 66.2 & 65.1     & 63.8  & - & - & - \\ 
\quad\quad + MUPPET        & \textbf{83.8}         &   \textbf{69.4}        &  \textbf{69.0}                    & \textbf{64.6}  & - & - & - \\ \\
RoBERTa-L & 86.4   & 78.1  & 83.4         &     73.6          & - & - & - \\
\quad\quad + MUPPET        & \textbf{\underline{87.5}}         &  \textbf{\underline{79.2}}    &   \textbf{\underline{86.4}}        &    \textbf{74.4}   & - & - & - \\ 
\\
BART & 86.2       & \textbf{78.1}     & \textbf{84.1}      & \textbf{71.4}   & 44.16/21.28/40.90 & 39.29/20.09/35.65 & 24.19/8.12/21.31       \\
\quad\quad + MUPPET        & \textbf{86.9}         &  74.8    &   75.9        &    70.8   & \textbf{\underline{44.45}/21.25/\underline{41.4}} & \textbf{\underline{40.40}/\underline{20.54}/36.21} & \textbf{\underline{30.30}/\underline{11.25}/\underline{24.92}}\\ \midrule
T5-L & 86.2 & 75.6 & 83.9 & 70.4 & 42.50/20.68/39.75 & - & -\\
T5-11B & 86.8 & 78.9 & 85.8 & \underline{75.4} & 43.52/\underline{21.55}/40.69 & - & -\\
PEGASUS & - & - & - & -  & 44.17/\textbf{21.47}/41.11 & 39.12/19.86/36.24 & 26.63/9.01/21.60\\
ERNIE-GEN & - & - & - & - & 44.02/\textbf{21.17}/\textbf{41.26} & 39.25/ 20.25/\textbf{36.53} & -\\
ProphetNet & - & - & - & - & 44.20/21.17/\textbf{41.30} & 39.51/20.42/\textbf{\underline{36.69}} & - \\
\bottomrule
\end{tabular}
\caption{We present results for the non-GLUE Sentence Prediction tasks as well as a set of standard Commonsense tasks. Bolded numbers signify MUPPET vs. base model, while an underline signifies the best number. If not explicitly stated, the results are showing the accuracy of the evaluation set. For commonsense tasks, we re-use the task head from pre-finetuning.}
\label{table:mtl_sp_commonsense}
\end{table*}

\begin{table*}[th]
\centering
\small
\begin{tabular}{@{}lllll|llll@{}}
\toprule
 & \multicolumn{1}{c}{SP} & \multicolumn{3}{c}{Structured Prediction (Penn)} & \multicolumn{3}{c}{Summarization}\\
 \cmidrule(lr){2-2}\cmidrule(lr){3-5}\cmidrule(lr){6-8} \\
 & 
  Hyperpartisan &
  Chunking &
  Parsing &
  POS & 
  Arxiv & 
  PubMed      & 
  BigPatent\\ \midrule

RoBERTa-B   & 84.2       & 93.4 & \textbf{95.1}     & \textbf{93.7}  & - & - &-  \\
 \quad\quad + MUPPET        & \textbf{85.8}         &   \textbf{95.5}        &  94.5                    & 93.2  & - & - &- \\
RoBERTa-L & 90.4   &  95.1 & 94.5        &     93.4  & - & - &-         \\
\quad\quad + MUPPET        & \textbf{\underline{92.5}}         &  \textbf{\underline{96.9}}    &   \textbf{\underline{95.7}}        &    \textbf{\underline{97.9}}  & - & - &-	
   \\ \midrule
BART & 85.1       & {92.1}     & {91.1}      & {91.8}   & 41.20/9.20/32.45 & 39.87/16.43/35.56 & 48.54/29.35/39.42       \\
\quad\quad + MUPPET        & \textbf{87.2}       & \textbf{96.1}     & \textbf{94.5}      & \textbf{97.2}   & \textbf{\underline{43.90}}/\textbf{14.50}/\textbf{\underline{40.10}} & \textbf{\underline{45.13}}/\textbf{\underline{19.80}}/\textbf{39.90} & \textbf{\underline{52.34}}/\textbf{\underline{33.50}}/\textbf{\underline{42.80}}       \\ \midrule
Pegasus 
& - & - & - & - & 43.85/\underline{16.83}/39.17 & 44.53/19.30/\underline{40.70} & 52.25/33.04/41.80 \\
\bottomrule
\end{tabular}
\caption{We present results on a large set of different tasks across datasets that are not available to the model during the pre-finetuning stage. Bolded numbers signify MUPPET vs. base model, while an underline signifies the best number. For Chunking/Parsing, we use F1, while for Part-Of-Speech tagging, we use accuracy.}
\label{table:mtl_out_of_mtl}
\end{table*}

\begin{table*}[t]
    \centering
    \small
    \begin{tabular}{llrrrr}
    \toprule
    \textbf{Model} & \textbf{Training Data} & A1 & A2 & A3 & ANLI \\ \midrule
 RoBERTa & S,M & 47.6 & 25.4 & 22.1 & 31.1 \\
 	& +F & 54.0 & 24.2 & 22.4 & 32.8 \\
 	& +F+A1\textbf{$^{\star 2}$} & 68.7 & 19.3 & 22.0 & 35.8 \\
 	& +F+A1+A2\textbf{$^{\star 3}$} & 71.2 & 44.3 & 20.4 & 43.7 
 	 \\
 	& S,M,F,ANLI & 73.8 & 48.9 & \textbf{44.4} & 53.7 \\ 
 	\\ 
 RoBERTa-MUPPET & S,M & \textbf{49.9} & \textbf{28.2} & \textbf{24.2} & \textbf{33.3} \\
 	& +F & \textbf{55.2} & \textbf{26.8} & \textbf{24.6} & \textbf{33.9} \\
 	& +F+A1\textbf{$^{\star 2}$} & \textbf{70.9} & \textbf{22.5} & \textbf{25.1} & \textbf{36.7} \\
 	& +F+A1+A2\textbf{$^{\star 3}$} & \textbf{74.3} & \textbf{48.2} & \textbf{22.8} & \textbf{45.9} 
 	 \\
 	& S,M,F,ANLI & \textbf{\underline{76.9}} & \textbf{52.3} & 44.2 & \textbf{56.9} \\\midrule
InfoBERT \cite{infobert} 	& S,M,F,ANLI & 76.4 & 51.6 & \underline{48.6} & \underline{58.3} \\
ALUM \cite{ALUM}	& S,M,F,ANLI & 73.3 & \underline{53.4} & 48.2 & 57.7 \\
XL-NET \cite{XLNET} 	& S,M,F,ANLI & 67.6 & 50.7 & 48.3 & 55.1 \\
 	\bottomrule
 	 \end{tabular}
     \vspace{-5pt}
     \caption{We show the performance of the RoBERTa model and the pre-finetuned RoBERTa-MUPPET model on the ANLI benchmark. Bolded numbers signify MUPPET vs base model, underline signifies best number. `S' refers to \textsc{SNLI}, `M' to \textsc{MNLI} dev (-m=matched, -mm=mismatched), and `F' to \textsc{FEVER}; `A1--A3' refer to the rounds respectively and `ANLI' refers to A1+A2+A3.}
     \label{table:anli_perf}
\end{table*}
Across the board, pre-trained representations that were further refined with pre-finetuning outperformed standard pre-trained representations. We see more modest gains on larger datasets, most likely because we do not need to refine representations beforehand if the fine-tuning dataset is large. On smaller datasets, we see substantial gains. For example, the pre-finetuned RoBERTa-BASE model on RTE improves by close to 9 points, rivaling the RoBERTa-Large accuracy, while the pre-finetuned RoBERTa-Large model gets new state-of-the-art on RTE rivaling models an order of magnitude larger than it.

We do not improve just over sentence prediction tasks but on every set of tasks that we measured. For example, we reach a new state of the art on the HellaSwag dataset previously achieved by utilizing a new fine-tuning approach. Our methods do not increase parameter count or any complexity measures but are quite successful at refining features and preparing them for downstream fine-tuning.

\subsection{Finetuning Outside of Pre-Finetuning Domain}
We also report the performance on tasks not included in the pre-finetuning data. To do so, we finetune our models on a set of tasks including (1) ANLI \cite{anli} and Hyperpartisan \cite{hyperpartisan} for classification, (2) Arxiv \cite{arxiv}, PubMed \cite{pubmed}, \cite{bigpatent} for summarization, and (3) Chunking, Constituency Parsing and Part-Of-Speech tagging for structured prediction from the Penn Treebank dataset \cite{penn}. We present these results in Table~\ref{table:mtl_out_of_mtl} and Table~\ref{table:anli_perf}.

We see that the MUPPET variants of our models out-perform the baselines consistently across task type and dataset. As a special case we do an in depth analysis of the MUPPET variant of RoBERTa on the notoriously tough ANLI dataset and see the same pattern. Pre-finetuned models consistently outperform their base counterparts.

\section{Understanding Multi-Task at Scale}
\label{sec:understanding_mtl}
\subsection{Importance of Scale}
\label{sec:scale_ablation}
The first axis we would like to explore is the scale on which multi-task learning is done. Previous work, such as T5 and MT-DNN, focused on the MTL scale of around a dozen datasets. To the best of our knowledge, our paper has the largest MTL set up to date. Accordingly, we are interested in empirically exploring the effects of scaling up the number of datasets to the representations learned during MTL.

We pre-finetune a collection of \textit{RoBERTa-Base} models with varying numbers of datasets. We train seven models, six uniformly chosen between 10 and 40, ensuring that at each point, the selected datasets are a superset of the datasets from prior points. The last model is fully trained on all datasets. Concretely given two models trained with a different number of datasets $a,b : a>b$, model $a$ will contain all datasets used to train model $b$ and more.

\begin{figure*}
    \centering
    \includegraphics[width=1.0\textwidth]{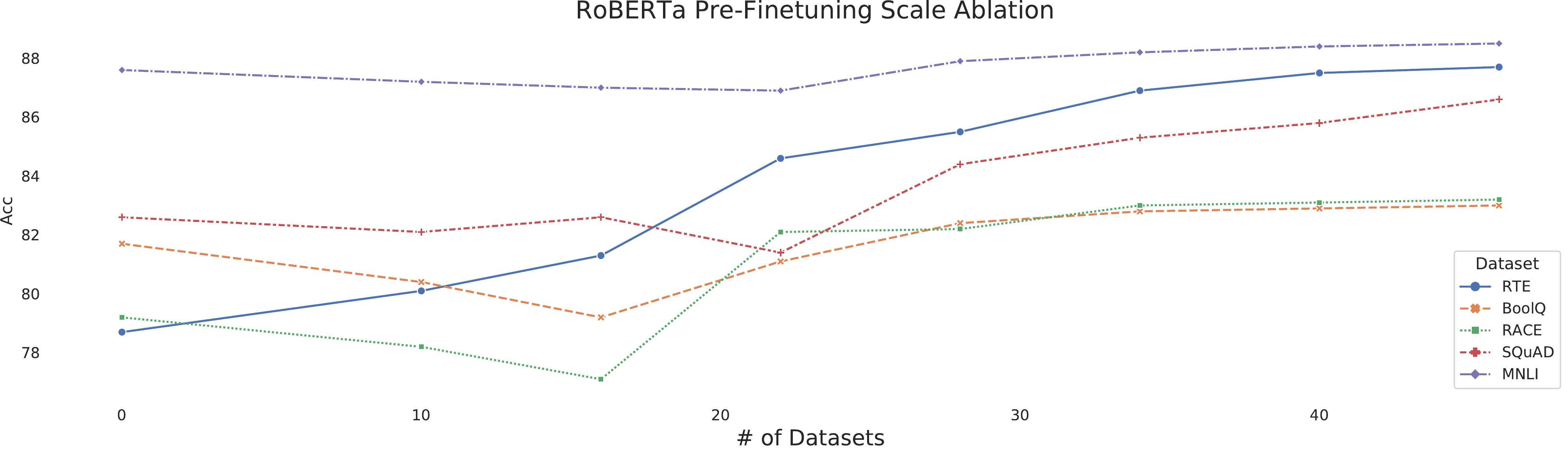}
    \caption{We plot the RoBERTa evaluation accuracy of five datasets: RTE, BoolQ, RACE, SQuAD, and MNLI, across various scales of multi-task learning measured in the number of datasets. We notice that performance initially degrades until a critical point is reached regarding the number of the datasets used by the MTL framework for all but one dataset. Post this critical point; our representations improve over the original RoBERTa model.}
    \label{fig:scale_ablation}
\end{figure*}

For each version of the model, we fine-tune five datasets and plot the results in Figure~\ref{fig:scale_ablation}. Specifically we finetune STS-B \citep{stsb}, BoolQ \citep{clark2019boolq}, RACE \citep{race}, SQuAD \citep{race}, and MNLI \citep{mnli}. We include these five datasets in the first MTL run (10 datasets) to remove any bias from adding them in a later stage.

We see a couple of interesting patterns. First, for individual tasks such as RTE \citep{rte}, increasing the pre-finetuning scale monotonically improves performance. This is aligned with other papers that have seen benefits from first training on MNLI \citep{mnli} and then fine-tuning on RTE \citep{ROBERTA}. For other datasets, we see that doing MTL in the $<15$ datasets regime is detrimental for end-task fine-tuning. This is also aligned with other empirical observations, i.e., T5 reported that doing MTL did not improve over only fine-tuning. Nevertheless, it seems that as we increase the number of tasks past some critical point, our pre-trained representations become more generalizable. Furthermore, although dependent on the dataset, this critical point is roughly between 10 and 25 tasks. 

This suggests that previously observed MTL limitations were not fundamental and can instead be attributed to the lack of sufficient scale.

\subsection{Importance of Heterogenous Batches}
\label{sec:batches}
Another critical factor to getting MTL to learn generalizable representations is the method through which MTL is implemented, specifically the selection of batches. To better quantify this trend, we experimented with three balancing schemes: dataset homogenous, batch homogenous and batch heterogenous.

We refer to dataset homogenous as selecting batches from datasets sequentially. So we first train on dataset $A$, then train on dataset $B$, etc. On the other hand, batch homogenous refers to selecting batches containing only data from the same task; therefore, all gradients are from the same dataset. This is implemented by selecting all datasets, batching on a dataset level, and selecting those same batches randomly during training. Finally, batch heterogeneous refers to a single update containing a batch from multiple different datasets spanning different tasks. We implemented this by first creating homogenous sub-batches, calculating loss per sub-batch per GPU, and then aggregating across GPUs manifesting in a gradient update that contains various datasets and, therefore, tasks.

To dissect the importance of heterogeneous batches, we train a RoBERTa-Base model on 35 randomly selected tasks using the three data selection methodologies outlined above. We then fine-tune these three models on the same five data-sets mentioned in the previous section.

\begin{figure*}
    \centering
    \includegraphics[width=1.0\textwidth]{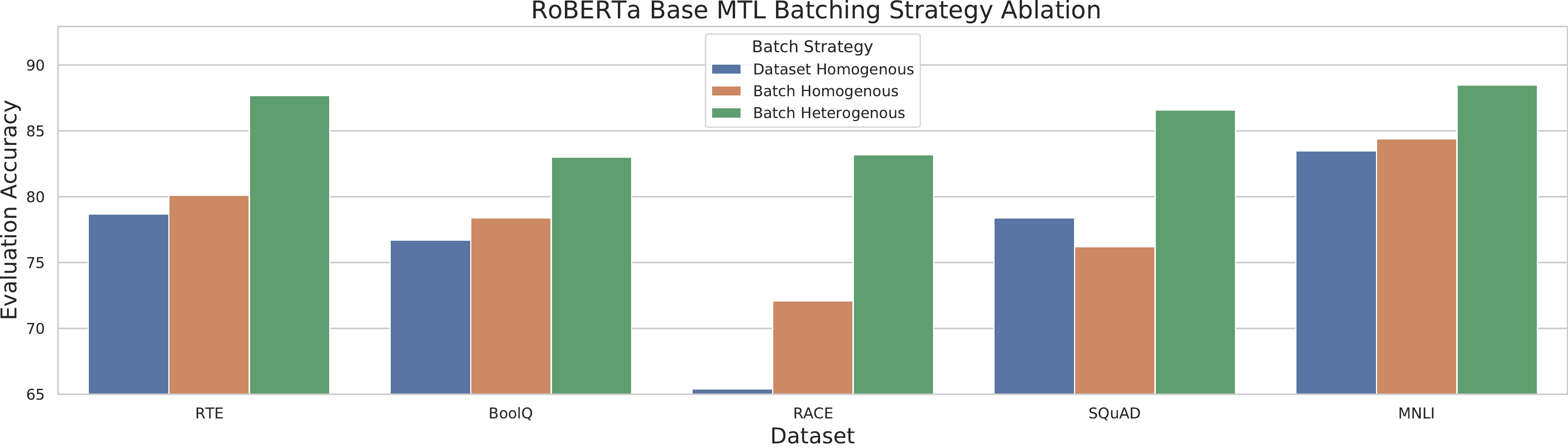}
    \caption{We plot the evaluation accuracy of RoBERTa across five datasets: RTE, BoolQ, RACE, SQuAD, and MNLI, using our three batching strategies for multi-task: Dataset Homogeneous, Batch Homogeneous, Batch Heterogeneous. The use of heterogenous batches outperforms other batching strategies by a significant margin and highlights the importance of implementing MTL with the correct batching strategy.}
    \label{fig:batch_strat_ablation}
\end{figure*}

We present our results in Figure~\ref{fig:batch_strat_ablation}. We see the importance of properly defining a batching strategy for effective multi-task learning. Our findings are also consistent with \cite{RXF} which saw that sequential training of data-sets degrades generalizable representations.

\subsection{Low Resource Experiments}
We noticed in Section~\S\ref{sec:empirical_results} that data-sets with smaller data-set sizes tended to improve more from MTL training. To strengthen this hypothesis, we look at two factors: the scale of pre-finetuning and the scale of fine-tuning (size of fine-tuning data-set).

We select three data-sets that were not used in pre-finetuning in Section~\S\ref{sec:scale_ablation}. We also select nine partitions per fine-tuning data-set, which is sampled uniformly between 10\% of the data-set and 100\% of the data-set. Selecting the low-resource splits was done through random sampling.

\begin{figure*}[h]
\centering
\includegraphics[width=.48\textwidth]{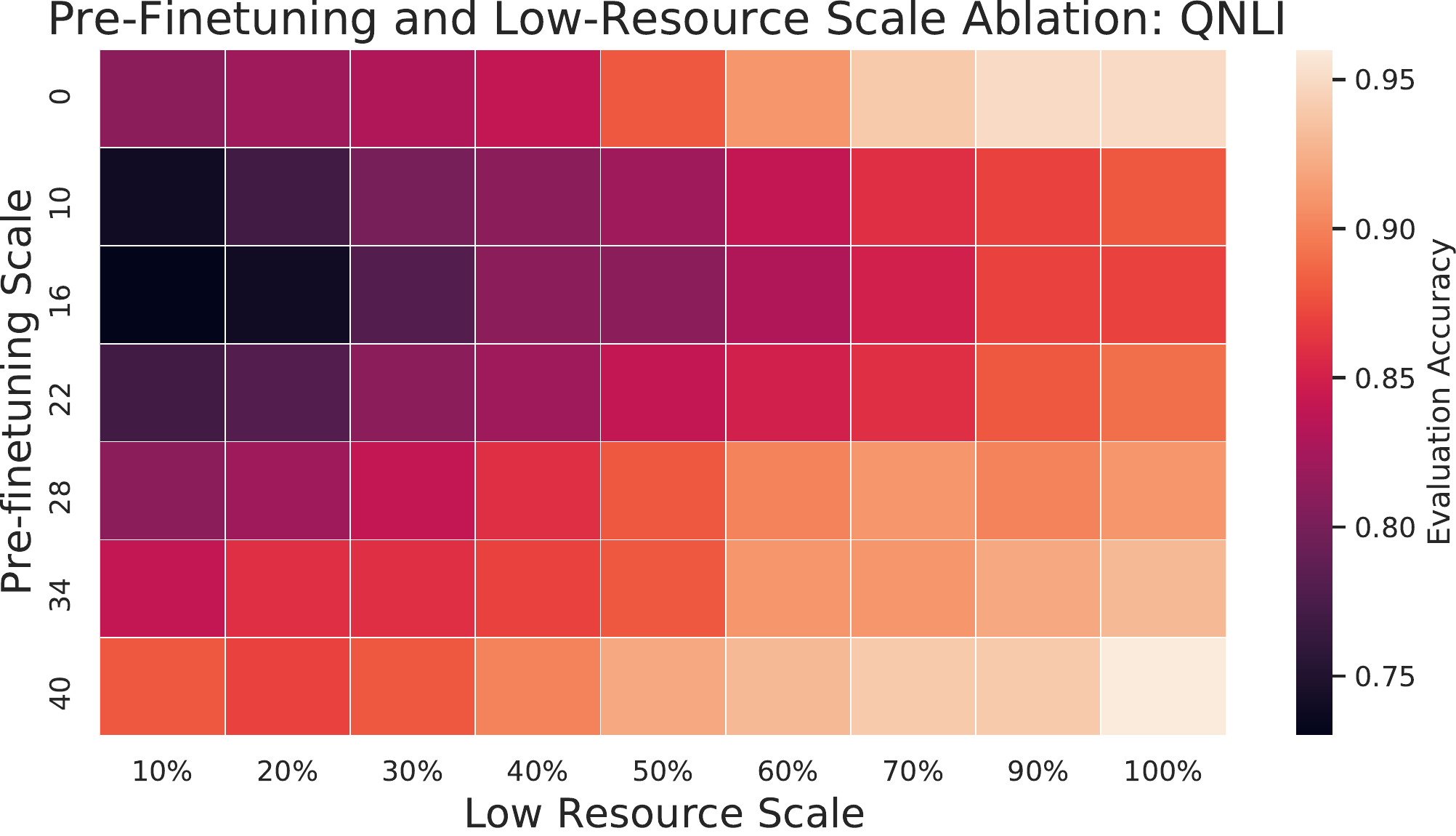}\hfill
\includegraphics[width=.48\textwidth]{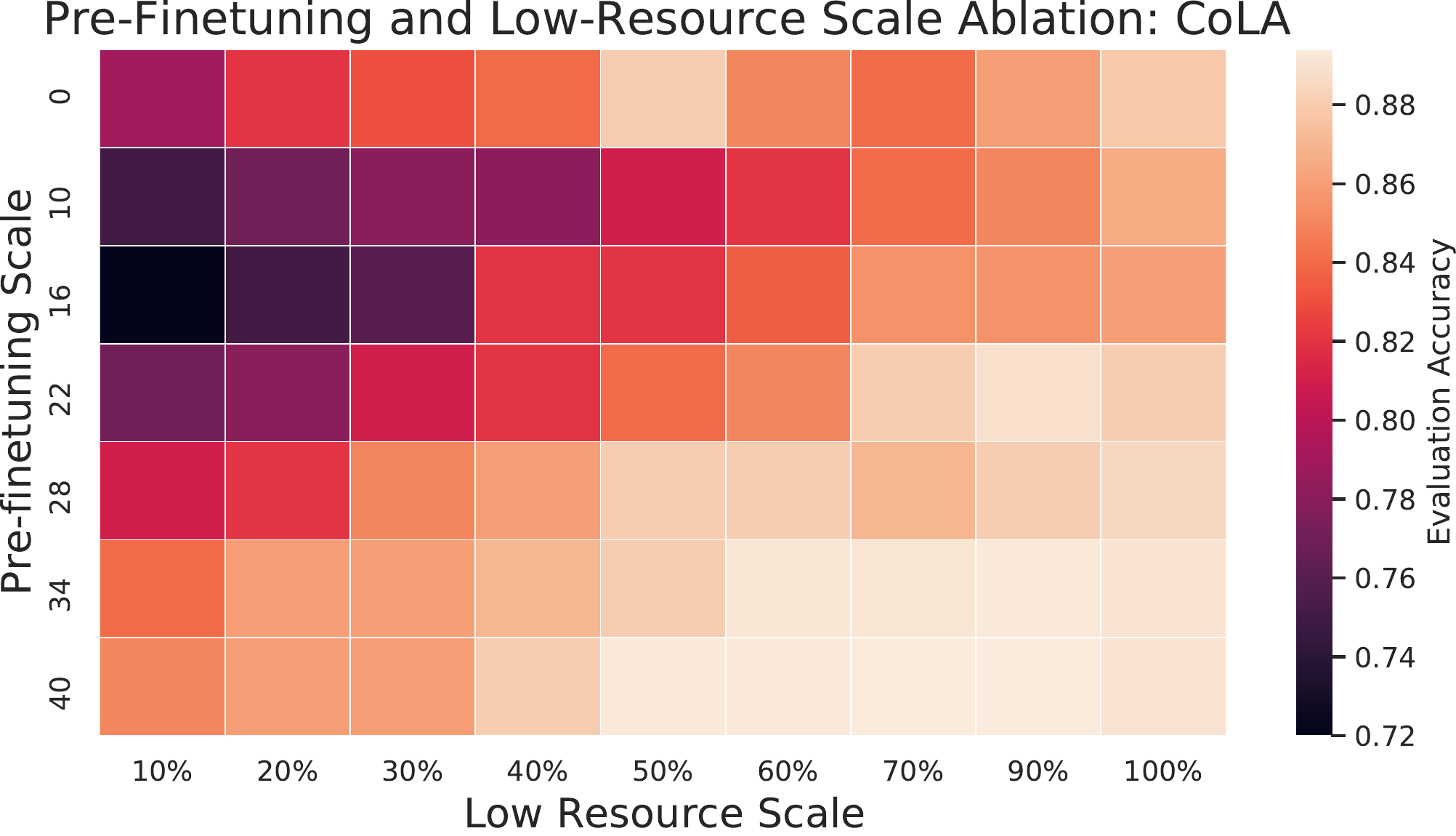}
\caption{We fine-tune every low-resource split with every pre-finetuning
checkpoint from Section~\S\ref{sec:scale_ablation} for two datasets not available in any of the pre-finetuning MTL datasets; QNLI \citep{qnli} and CoLA \citep{cola}. The pre-finetuning scale is reported in terms of the number of datasets.}
\label{fig:low_resource_scale_abl}
\end{figure*}

We then fine-tune every low-resource split with every pre-finetuning checkpoint from Section~\S\ref{sec:scale_ablation}. We plot the heatmaps generated from these runs in Figure~\ref{fig:low_resource_scale_abl}.

Multiple patterns emerge. First, we see a clear visualization of the critical point mentioned when doing pre-finetuning. As we increase the scale of MTL, better representations are available for downstream finetuning. Furthermore, we see that pre-finetuned models at a larger scale are much more data-efficient than standard pre-trained models.

Specifically looking at the 34/40 pre-finetuning scale on Figure~\ref{fig:low_resource_scale_abl} we see that we reach higher evaluation accuracies much sooner than the base RoBERTa model (row 0).

\section{Conclusion}
In this work, we propose \textit{pre-finetuning}, a stage after pre-training to further refine representations before end-task finetuning. We show that we can effectively learn more robust representations through multi-task learning (MTL) at scale. Our MTL models outperform their vanilla pre-trained counterparts across several tasks. Our analysis shows that properly scaling MTL with heterogeneous batches and loss scaling is critical to leveraging better representations. We also show a critical point regarding the number of tasks when doing multi-task learning, where fewer tasks degrade representations compared to the pre-trained model, but more tasks than this point improve representations. 

We discussed a practical setting in which doing this massive multi-task learning is stable and effective through simple loss scaling and heterogeneous batches. With our method, we improve upon prior state of the art methods for RTE \cite{rte} and HellaSWAG \cite{hellaswag}, as well as improve upon vanilla pre-trained representations for MNLI \cite{mnli}, SQuAD \cite{SQUAD}, BoolQ \cite{clark2019boolq}, and Common Sense QA \cite{commonsenseqa}. We also our MTL model performance with low resource experiments. We show that on held-out datasets, leveraging representations from our pre-finetuned models with 34-40 tasks, we reach higher evaluation accuracies with much less data than the RoBERTa model.
 
\clearpage

\bibliography{anthology,acl2020}

\begin{thebibliography}{73}
\expandafter\ifx\csname natexlab\endcsname\relax\def\natexlab#1{#1}\fi

\bibitem[{Aghajanyan et~al.(2020)Aghajanyan, Shrivastava, Gupta, Goyal,
  Zettlemoyer, and Gupta}]{RXF}
Armen Aghajanyan, Akshat Shrivastava, Anchit Gupta, Naman Goyal, Luke
  Zettlemoyer, and Sonal Gupta. 2020.
\newblock Better fine-tuning by reducing representational collapse.
\newblock \emph{arXiv preprint arXiv:2008.03156}.

\bibitem[{Amini et~al.(2019)Amini, Gabriel, Lin, Koncel-Kedziorski, Choi, and
  Hajishirzi}]{mathqa}
Aida Amini, Saadia Gabriel, Peter Lin, Rik Koncel-Kedziorski, Yejin Choi, and
  Hannaneh Hajishirzi. 2019.
\newblock Mathqa: Towards interpretable math word problem solving with
  operation-based formalisms.
\newblock \emph{arXiv preprint arXiv:1905.13319}.

\bibitem[{Bentivogli et~al.(2009)Bentivogli, Clark, Dagan, and
  Giampiccolo}]{rte}
Luisa Bentivogli, Peter Clark, Ido Dagan, and Danilo Giampiccolo. 2009.
\newblock The fifth pascal recognizing textual entailment challenge.
\newblock In \emph{TAC}.

\bibitem[{Bowman et~al.(2015)Bowman, Angeli, Potts, and
  Manning}]{snli:emnlp2015}
Samuel Bowman, Gabor Angeli, Christopher Potts, and Christopher Manning. 2015.
\newblock A large annotated corpus for learning natural language inference.
\newblock In \emph{Proceedings of the 2015 Conference on Empirical Methods in
  Natural Language Processing (EMNLP)}. Association for Computational
  Linguistics.

\bibitem[{Cer et~al.(2017)Cer, Diab, Agirre, Lopez-Gazpio, and Specia}]{stsb}
Daniel Cer, Mona Diab, Eneko Agirre, Inigo Lopez-Gazpio, and Lucia Specia.
  2017.
\newblock Semeval-2017 task 1: Semantic textual similarity-multilingual and
  cross-lingual focused evaluation.
\newblock \emph{arXiv preprint arXiv:1708.00055}.

\bibitem[{Chen et~al.(2018)Chen, Badrinarayanan, Lee, and
  Rabinovich}]{gradnorm_mtl}
Zhao Chen, Vijay Badrinarayanan, Chen-Yu Lee, and Andrew Rabinovich. 2018.
\newblock Gradnorm: Gradient normalization for adaptive loss balancing in deep
  multitask networks.
\newblock In \emph{International Conference on Machine Learning}, pages
  794--803. PMLR.

\bibitem[{Clark et~al.(2019)Clark, Lee, Chang, Kwiatkowski, Collins, and
  Toutanova}]{clark2019boolq}
Christopher Clark, Kenton Lee, Ming-Wei Chang, Tom Kwiatkowski, Michael
  Collins, and Kristina Toutanova. 2019.
\newblock {B}ool{Q}: Exploring the surprising difficulty of natural yes/no
  questions.
\newblock In \emph{Proceedings of NAACL-HLT 2019}.

\bibitem[{Clark et~al.(2020)Clark, Luong, Le, and Manning}]{ELECTRA}
Kevin Clark, Minh-Thang Luong, Quoc~V Le, and Christopher~D Manning. 2020.
\newblock Electra: Pre-training text encoders as discriminators rather than
  generators.
\newblock \emph{arXiv preprint arXiv:2003.10555}.

\bibitem[{Clark et~al.(2018)Clark, Cowhey, Etzioni, Khot, Sabharwal, Schoenick,
  and Tafjord}]{arc}
Peter Clark, Isaac Cowhey, Oren Etzioni, Tushar Khot, Ashish Sabharwal, Carissa
  Schoenick, and Oyvind Tafjord. 2018.
\newblock Think you have solved question answering? try arc, the ai2 reasoning
  challenge.
\newblock \emph{arXiv preprint arXiv:1803.05457}.

\bibitem[{Cohan et~al.(2018)Cohan, Dernoncourt, Kim, Bui, Kim, Chang, and
  Goharian}]{pubmed}
Arman Cohan, Franck Dernoncourt, Doo~Soon Kim, Trung Bui, Seokhwan Kim, Walter
  Chang, and Nazli Goharian. 2018.
\newblock A discourse-aware attention model for abstractive summarization of
  long documents.
\newblock \emph{arXiv preprint arXiv:1804.05685}.

\bibitem[{De~Marneffe et~al.(2019)De~Marneffe, Simons, and
  Tonhauser}]{demarneffe:cb}
Marie-Catherine De~Marneffe, Mandy Simons, and Judith Tonhauser. 2019.
\newblock {The CommitmentBank}: Investigating projection in naturally occurring
  discourse.
\newblock To appear in proceedings of Sinn und Bedeutung 23. Data can be found
  at https://github.com/mcdm/CommitmentBank/.

\bibitem[{Devlin et~al.(2018)Devlin, Chang, Lee, and Toutanova}]{BERT}
Jacob Devlin, Ming-Wei Chang, Kenton Lee, and Kristina Toutanova. 2018.
\newblock Bert: Pre-training of deep bidirectional transformers for language
  understanding.
\newblock \emph{arXiv preprint arXiv:1810.04805}.

\bibitem[{DeYoung et~al.()DeYoung, Jain, Rajani, Lehman, Xiong, Socher, and
  Wallace}]{eraser}
Jay DeYoung, Sarthak Jain, Nazneen~Fatema Rajani, Eric Lehman, Caiming Xiong,
  Richard Socher, and Byron~C. Wallace.
\newblock Eraser: A benchmark to evaluate rationalized nlp models.

\bibitem[{Dolan and Brockett(2005)}]{mrpc}
William~B Dolan and Chris Brockett. 2005.
\newblock Automatically constructing a corpus of sentential paraphrases.
\newblock In \emph{Proceedings of the Third International Workshop on
  Paraphrasing (IWP2005)}.

\bibitem[{Dua et~al.(2019)Dua, Wang, Dasigi, Stanovsky, Singh, and
  Gardner}]{DROP}
Dheeru Dua, Yizhong Wang, Pradeep Dasigi, Gabriel Stanovsky, Sameer Singh, and
  Matt Gardner. 2019.
\newblock Drop: A reading comprehension benchmark requiring discrete reasoning
  over paragraphs.
\newblock In \emph{Proc. of NAACL}.

\bibitem[{Eidelman(2019)}]{billsum}
Vladimir Eidelman. 2019.
\newblock Billsum: A corpus for automatic summarization of us legislation.
\newblock In \emph{Proceedings of the 2nd Workshop on New Frontiers in
  Summarization}, pages 48--56.

\bibitem[{Fabbri et~al.(2019)Fabbri, Li, She, Li, and Radev}]{multinews}
Alexander~R Fabbri, Irene Li, Tianwei She, Suyi Li, and Dragomir~R Radev. 2019.
\newblock Multi-news: A large-scale multi-document summarization dataset and
  abstractive hierarchical model.
\newblock \emph{arXiv preprint arXiv:1906.01749}.

\bibitem[{He et~al.(2019)He, Wang, Liu, Feng, and Wu}]{arxiv}
Jun He, Liqun Wang, Liu Liu, Jiao Feng, and Hao Wu. 2019.
\newblock Long document classification from local word glimpses via recurrent
  attention learning.
\newblock \emph{IEEE Access}, 7:40707--40718.

\bibitem[{Hermann et~al.(2015)Hermann, Kocisky, Grefenstette, Espeholt, Kay,
  Suleyman, and Blunsom}]{cnndailymail}
Karl~Moritz Hermann, Tomas Kocisky, Edward Grefenstette, Lasse Espeholt, Will
  Kay, Mustafa Suleyman, and Phil Blunsom. 2015.
\newblock Teaching machines to read and comprehend.
\newblock In \emph{Advances in neural information processing systems}, pages
  1693--1701.

\bibitem[{Hovy et~al.(2001)Hovy, Gerber, Hermjakob, Lin, and
  Ravichandran}]{hovy-etal-2001-toward}
Eduard Hovy, Laurie Gerber, Ulf Hermjakob, Chin-Yew Lin, and Deepak
  Ravichandran. 2001.
\newblock \href {https://www.aclweb.org/anthology/H01-1069} {Toward
  semantics-based answer pinpointing}.
\newblock In \emph{Proceedings of the First International Conference on Human
  Language Technology Research}.

\bibitem[{Huang et~al.(2019)Huang, Bras, Bhagavatula, and Choi}]{cosmosqa}
Lifu Huang, Ronan~Le Bras, Chandra Bhagavatula, and Yejin Choi. 2019.
\newblock Cosmos qa: Machine reading comprehension with contextual commonsense
  reasoning.
\newblock \emph{arXiv preprint arXiv:1909.00277}.

\bibitem[{Iyer et~al.(2017)Iyer, Dandekar, and Csernai}]{qqp}
Shankar Iyer, Nikhil Dandekar, and Kornel Csernai. 2017.
\newblock \href
  {https://data.quora.com/First-Quora-Dataset-Release-Question-Pairs} {First
  quora dataset release: Question pairs}.

\bibitem[{{Joshi} et~al.(2017){Joshi}, {Choi}, {Weld}, and
  {Zettlemoyer}}]{TriviaQA}
Mandar {Joshi}, Eunsol {Choi}, Daniel {Weld}, and Luke {Zettlemoyer}. 2017.
\newblock \href {http://arxiv.org/abs/1705.03551} {{triviaqa: A Large Scale
  Distantly Supervised Challenge Dataset for Reading Comprehension}}.
\newblock \emph{arXiv e-prints}, page arXiv:1705.03551.

\bibitem[{Khashabi et~al.(2018)Khashabi, Chaturvedi, Roth, Upadhyay, and
  Roth}]{khashabi2018looking}
Daniel Khashabi, Snigdha Chaturvedi, Michael Roth, Shyam Upadhyay, and Dan
  Roth. 2018.
\newblock Looking beyond the surface: A challenge set for reading comprehension
  over multiple sentences.
\newblock In \emph{Proceedings of the 2018 Conference of the North American
  Chapter of the Association for Computational Linguistics: Human Language
  Technologies, Volume 1 (Long Papers)}, pages 252--262.

\bibitem[{Khashabi et~al.(2020)Khashabi, Min, Khot, Sabharwal, Tafjord, Clark,
  and Hajishirzi}]{unifiedqa}
Daniel Khashabi, Sewon Min, Tushar Khot, Ashish Sabharwal, Oyvind Tafjord,
  Peter Clark, and Hannaneh Hajishirzi. 2020.
\newblock \href {http://arxiv.org/abs/2005.00700} {Unifiedqa: Crossing format
  boundaries with a single qa system}.

\bibitem[{Khot et~al.(2018)Khot, Sabharwal, and Clark}]{scitail}
Tushar Khot, Ashish Sabharwal, and Peter Clark. 2018.
\newblock {SciTail}: A textual entailment dataset from science question
  answering.
\newblock In \emph{AAAI}.

\bibitem[{Kiesel et~al.(2019)Kiesel, Mestre, Shukla, Vincent, Adineh, Corney,
  Stein, and Potthast}]{hyperpartisan}
Johannes Kiesel, Maria Mestre, Rishabh Shukla, Emmanuel Vincent, Payam Adineh,
  David Corney, Benno Stein, and Martin Potthast. 2019.
\newblock Semeval-2019 task 4: Hyperpartisan news detection.
\newblock In \emph{Proceedings of the 13th International Workshop on Semantic
  Evaluation}, pages 829--839.

\bibitem[{Kingma and Ba(2014)}]{ADAM}
Diederik~P Kingma and Jimmy Ba. 2014.
\newblock Adam: A method for stochastic optimization.
\newblock \emph{arXiv preprint arXiv:1412.6980}.

\bibitem[{Kwiatkowski et~al.(2019)Kwiatkowski, Palomaki, Redfield, Collins,
  Parikh, Alberti, Epstein, Polosukhin, Kelcey, Devlin, Lee, Toutanova, Jones,
  Chang, Dai, Uszkoreit, Le, and Petrov}]{NaturalQuestions}
Tom Kwiatkowski, Jennimaria Palomaki, Olivia Redfield, Michael Collins, Ankur
  Parikh, Chris Alberti, Danielle Epstein, Illia Polosukhin, Matthew Kelcey,
  Jacob Devlin, Kenton Lee, Kristina~N. Toutanova, Llion Jones, Ming-Wei Chang,
  Andrew Dai, Jakob Uszkoreit, Quoc Le, and Slav Petrov. 2019.
\newblock Natural questions: a benchmark for question answering research.
\newblock \emph{Transactions of the Association of Computational Linguistics}.

\bibitem[{Lai et~al.(2017)Lai, Xie, Liu, Yang, and Hovy}]{race}
Guokun Lai, Qizhe Xie, Hanxiao Liu, Yiming Yang, and Eduard Hovy. 2017.
\newblock Race: Large-scale reading comprehension dataset from examinations.
\newblock \emph{arXiv preprint arXiv:1704.04683}.

\bibitem[{Levesque et~al.(2012)Levesque, Davis, and Morgenstern}]{winograd}
Hector Levesque, Ernest Davis, and Leora Morgenstern. 2012.
\newblock The winograd schema challenge.
\newblock In \emph{Thirteenth International Conference on the Principles of
  Knowledge Representation and Reasoning}. Citeseer.

\bibitem[{Levesque et~al.(2011)Levesque, Davis, and
  Morgenstern}]{levesque2011winograd}
Hector~J Levesque, Ernest Davis, and Leora Morgenstern. 2011.
\newblock The {W}inograd schema challenge.
\newblock In \emph{{AAAI} Spring Symposium: Logical Formalizations of
  Commonsense Reasoning}, volume~46, page~47.

\bibitem[{Lewis et~al.(2019)Lewis, Liu, Goyal, Ghazvininejad, Mohamed, Levy,
  Stoyanov, and Zettlemoyer}]{BART}
Mike Lewis, Yinhan Liu, Naman Goyal, Marjan Ghazvininejad, Abdelrahman Mohamed,
  Omer Levy, Ves Stoyanov, and Luke Zettlemoyer. 2019.
\newblock Bart: Denoising sequence-to-sequence pre-training for natural
  language generation, translation, and comprehension.
\newblock \emph{arXiv preprint arXiv:1910.13461}.

\bibitem[{Li and Roth(2002)}]{li-roth-2002-learning}
Xin Li and Dan Roth. 2002.
\newblock \href {https://www.aclweb.org/anthology/C02-1150} {Learning question
  classifiers}.
\newblock In \emph{{COLING} 2002: The 19th International Conference on
  Computational Linguistics}.

\bibitem[{Liu et~al.(2020)Liu, Cheng, He, Chen, Wang, Poon, and Gao}]{ALUM}
X.~Liu, Hao Cheng, Pengcheng He, W.~Chen, Yu~Wang, Hoifung Poon, and Jianfeng
  Gao. 2020.
\newblock Adversarial training for large neural language models.
\newblock \emph{ArXiv}, abs/2004.08994.

\bibitem[{Liu et~al.(2019{\natexlab{a}})Liu, He, Chen, and Gao}]{MT_DNN}
Xiaodong Liu, Pengcheng He, Weizhu Chen, and Jianfeng Gao. 2019{\natexlab{a}}.
\newblock Multi-task deep neural networks for natural language understanding.
\newblock \emph{arXiv preprint arXiv:1901.11504}.

\bibitem[{Liu et~al.(2019{\natexlab{b}})Liu, Ott, Goyal, Du, Joshi, Chen, Levy,
  Lewis, Zettlemoyer, and Stoyanov}]{ROBERTA}
Yinhan Liu, Myle Ott, Naman Goyal, Jingfei Du, Mandar Joshi, Danqi Chen, Omer
  Levy, Mike Lewis, Luke Zettlemoyer, and Veselin Stoyanov. 2019{\natexlab{b}}.
\newblock Roberta: A robustly optimized bert pretraining approach.
\newblock \emph{arXiv preprint arXiv:1907.11692}.

\bibitem[{Maas et~al.(2011)Maas, Daly, Pham, Huang, Ng, and Potts}]{IMDB}
Andrew~L. Maas, Raymond~E. Daly, Peter~T. Pham, Dan Huang, Andrew~Y. Ng, and
  Christopher Potts. 2011.
\newblock \href {http://www.aclweb.org/anthology/P11-1015} {Learning word
  vectors for sentiment analysis}.
\newblock In \emph{Proceedings of the 49th Annual Meeting of the Association
  for Computational Linguistics: Human Language Technologies}, pages 142--150,
  Portland, Oregon, USA. Association for Computational Linguistics.

\bibitem[{Marcus et~al.(1993)Marcus, Santorini, and Marcinkiewicz}]{penn}
Mitchell Marcus, Beatrice Santorini, and Mary~Ann Marcinkiewicz. 1993.
\newblock Building a large annotated corpus of english: The penn treebank.

\bibitem[{McCoy et~al.(2019)McCoy, Pavlick, and Linzen}]{HANS}
R.~Thomas McCoy, Ellie Pavlick, and Tal Linzen. 2019.
\newblock \href {http://arxiv.org/abs/1902.01007} {Right for the wrong reasons:
  Diagnosing syntactic heuristics in natural language inference}.
\newblock \emph{CoRR}, abs/1902.01007.

\bibitem[{Mihaylov et~al.(2018)Mihaylov, Clark, Khot, and
  Sabharwal}]{openbookqa}
Todor Mihaylov, Peter Clark, Tushar Khot, and Ashish Sabharwal. 2018.
\newblock Can a suit of armor conduct electricity? a new dataset for open book
  question answering.
\newblock \emph{arXiv preprint arXiv:1809.02789}.

\bibitem[{Mosbach et~al.(2020)Mosbach, Andriushchenko, and
  Klakow}]{stability_bert}
Marius Mosbach, Maksym Andriushchenko, and Dietrich Klakow. 2020.
\newblock On the stability of fine-tuning bert: Misconceptions, explanations,
  and strong baselines.
\newblock \emph{arXiv preprint arXiv:2006.04884}.

\bibitem[{Narayan et~al.(2018)Narayan, Cohen, and Lapata}]{xsum}
Shashi Narayan, Shay~B Cohen, and Mirella Lapata. 2018.
\newblock Don't give me the details, just the summary! topic-aware
  convolutional neural networks for extreme summarization.
\newblock \emph{arXiv preprint arXiv:1808.08745}.

\bibitem[{Nie et~al.(2019)Nie, Williams, Dinan, Bansal, Weston, and
  Kiela}]{anli}
Yixin Nie, Adina Williams, Emily Dinan, Mohit Bansal, Jason Weston, and Douwe
  Kiela. 2019.
\newblock Adversarial nli: A new benchmark for natural language understanding.
\newblock \emph{arXiv preprint arXiv:1910.14599}.

\bibitem[{Pang and Lee(2005)}]{RottenTomatoes}
Bo~Pang and Lillian Lee. 2005.
\newblock Seeing stars: Exploiting class relationships for sentiment
  categorization with respect to rating scales.
\newblock In \emph{Proceedings of the ACL}.

\bibitem[{Pilehvar and Camacho-Collados(2019)}]{pilehvar2018wic}
Mohammad~Taher Pilehvar and Jose Camacho-Collados. 2019.
\newblock {WiC}: The word-in-context dataset for evaluating context-sensitive
  meaning representations.
\newblock In \emph{Proceedings of NAACL-HLT}.

\bibitem[{Radford et~al.(2019)Radford, Wu, Child, Luan, Amodei, and
  Sutskever}]{GPT}
Alec Radford, Jeffrey Wu, Rewon Child, David Luan, Dario Amodei, and Ilya
  Sutskever. 2019.
\newblock Language models are unsupervised multitask learners.
\newblock \emph{OpenAI Blog}, 1(8):9.

\bibitem[{Raffel et~al.(2019)Raffel, Shazeer, Roberts, Lee, Narang, Matena,
  Zhou, Li, and Liu}]{T5}
Colin Raffel, Noam Shazeer, Adam Roberts, Katherine Lee, Sharan Narang, Michael
  Matena, Yanqi Zhou, Wei Li, and Peter~J Liu. 2019.
\newblock Exploring the limits of transfer learning with a unified text-to-text
  transformer.
\newblock \emph{arXiv preprint arXiv:1910.10683}.

\bibitem[{Rajpurkar et~al.(2016{\natexlab{a}})Rajpurkar, Zhang, Lopyrev, and
  Liang}]{SQUAD}
Pranav Rajpurkar, Jian Zhang, Konstantin Lopyrev, and Percy Liang.
  2016{\natexlab{a}}.
\newblock Squad: 100,000+ questions for machine comprehension of text.
\newblock \emph{arXiv preprint arXiv:1606.05250}.

\bibitem[{Rajpurkar et~al.(2016{\natexlab{b}})Rajpurkar, Zhang, Lopyrev, and
  Liang}]{qnli}
Pranav Rajpurkar, Jian Zhang, Konstantin Lopyrev, and Percy Liang.
  2016{\natexlab{b}}.
\newblock Squad: 100,000+ questions for machine comprehension of text.
\newblock \emph{arXiv preprint arXiv:1606.05250}.

\bibitem[{Roemmele et~al.(2011)Roemmele, Bejan, and
  Gordon}]{roemmele2011choice}
Melissa Roemmele, Cosmin~Adrian Bejan, and Andrew~S. Gordon. 2011.
\newblock Choice of plausible alternatives: An evaluation of commonsense causal
  reasoning.
\newblock In \emph{2011 AAAI Spring Symposium Series}.

\bibitem[{Seo et~al.(2016)Seo, Kembhavi, Farhadi, and Hajishirzi}]{bidaf}
Minjoon Seo, Aniruddha Kembhavi, Ali Farhadi, and Hannaneh Hajishirzi. 2016.
\newblock Bidirectional attention flow for machine comprehension.
\newblock \emph{arXiv preprint arXiv:1611.01603}.

\bibitem[{Sharma et~al.(2019)Sharma, Li, and Wang}]{bigpatent}
Eva Sharma, Chen Li, and Lu~Wang. 2019.
\newblock Bigpatent: A large-scale dataset for abstractive and coherent
  summarization.
\newblock \emph{arXiv preprint arXiv:1906.03741}.

\bibitem[{Socher et~al.(2013)Socher, Perelygin, Wu, Chuang, Manning, Ng, and
  Potts}]{sst2}
Richard Socher, Alex Perelygin, Jean Wu, Jason Chuang, Christopher~D Manning,
  Andrew Ng, and Christopher Potts. 2013.
\newblock Recursive deep models for semantic compositionality over a sentiment
  treebank.
\newblock In \emph{Proceedings of the 2013 conference on empirical methods in
  natural language processing}, pages 1631--1642.

\bibitem[{Srivastava et~al.(2014)Srivastava, Hinton, Krizhevsky, Sutskever, and
  Salakhutdinov}]{DROPOUT}
Nitish Srivastava, Geoffrey Hinton, Alex Krizhevsky, Ilya Sutskever, and Ruslan
  Salakhutdinov. 2014.
\newblock Dropout: a simple way to prevent neural networks from overfitting.
\newblock \emph{The journal of machine learning research}, 15(1):1929--1958.

\bibitem[{Szegedy et~al.(2015)Szegedy, Vanhoucke, Ioffe, Shlens, and
  Wojna}]{inception_label_smoothing}
Christian Szegedy, Vincent Vanhoucke, Sergey Ioffe, Jonathon Shlens, and
  Zbigniew Wojna. 2015.
\newblock Rethinking the inception architecture for computer vision. corr
  abs/1512.00567 (2015).

\bibitem[{Talmor et~al.(2018)Talmor, Herzig, Lourie, and
  Berant}]{commonsenseqa}
Alon Talmor, Jonathan Herzig, Nicholas Lourie, and Jonathan Berant. 2018.
\newblock Commonsenseqa: A question answering challenge targeting commonsense
  knowledge.
\newblock \emph{arXiv preprint arXiv:1811.00937}.

\bibitem[{Wang et~al.(2019)Wang, Pruksachatkun, Nangia, Singh, Michael, Hill,
  Levy, and Bowman}]{wang2019superglue}
Alex Wang, Yada Pruksachatkun, Nikita Nangia, Amanpreet Singh, Julian Michael,
  Felix Hill, Omer Levy, and Samuel~R. Bowman. 2019.
\newblock Super{GLUE}: A stickier benchmark for general-purpose language
  understanding systems.
\newblock \emph{arXiv preprint 1905.00537}.

\bibitem[{Wang et~al.(2018)Wang, Singh, Michael, Hill, Levy, and Bowman}]{GLUE}
Alex Wang, Amanpreet Singh, Julian Michael, Felix Hill, Omer Levy, and Samuel
  Bowman. 2018.
\newblock \href {https://doi.org/10.18653/v1/W18-5446} {{GLUE}: A multi-task
  benchmark and analysis platform for natural language understanding}.
\newblock In \emph{Proceedings of the 2018 {EMNLP} Workshop {B}lackbox{NLP}:
  Analyzing and Interpreting Neural Networks for {NLP}}, pages 353--355,
  Brussels, Belgium. Association for Computational Linguistics.

\bibitem[{Wang et~al.(2021)Wang, Wang, Cheng, Gan, Jia, Li, and Liu}]{infobert}
Boxin Wang, Shuohang Wang, Yu~Cheng, Zhe Gan, Ruoxi Jia, Bo~Li, and Jingjing
  Liu. 2021.
\newblock \href {https://openreview.net/forum?id=hpH98mK5Puk}
  {Info{\{}bert{\}}: Improving robustness of language models from an
  information theoretic perspective}.
\newblock In \emph{International Conference on Learning Representations}.

\bibitem[{Warstadt et~al.(2018)Warstadt, Singh, and Bowman}]{cola}
Alex Warstadt, Amanpreet Singh, and Samuel~R Bowman. 2018.
\newblock Neural network acceptability judgments.
\newblock \emph{arXiv preprint arXiv:1805.12471}.

\bibitem[{Welbl et~al.(2017)Welbl, Liu, and Gardner}]{sciq}
Johannes Welbl, Nelson~F Liu, and Matt Gardner. 2017.
\newblock Crowdsourcing multiple choice science questions.
\newblock \emph{arXiv preprint arXiv:1707.06209}.

\bibitem[{Williams et~al.(2018{\natexlab{a}})Williams, Nangia, and
  Bowman}]{mnli}
Adina Williams, Nikita Nangia, and Samuel Bowman. 2018{\natexlab{a}}.
\newblock \href {http://aclweb.org/anthology/N18-1101} {A broad-coverage
  challenge corpus for sentence understanding through inference}.
\newblock In \emph{Proceedings of the 2018 Conference of the North American
  Chapter of the Association for Computational Linguistics: Human Language
  Technologies, Volume 1 (Long Papers)}, pages 1112--1122. Association for
  Computational Linguistics.

\bibitem[{Williams et~al.(2018{\natexlab{b}})Williams, Nangia, and
  Bowman}]{MultiNLI}
Adina Williams, Nikita Nangia, and Samuel Bowman. 2018{\natexlab{b}}.
\newblock \href {http://aclweb.org/anthology/N18-1101} {A broad-coverage
  challenge corpus for sentence understanding through inference}.
\newblock In \emph{Proceedings of the 2018 Conference of the North American
  Chapter of the Association for Computational Linguistics: Human Language
  Technologies, Volume 1 (Long Papers)}, pages 1112--1122. Association for
  Computational Linguistics.

\bibitem[{Yang et~al.(2019)Yang, Dai, Yang, Carbonell, Salakhutdinov, and
  Le}]{XLNET}
Zhilin Yang, Zihang Dai, Yiming Yang, Jaime Carbonell, Russ~R Salakhutdinov,
  and Quoc~V Le. 2019.
\newblock Xlnet: Generalized autoregressive pretraining for language
  understanding.
\newblock In \emph{Advances in neural information processing systems}, pages
  5753--5763.

\bibitem[{Yang et~al.(2018)Yang, Qi, Zhang, Bengio, Cohen, Salakhutdinov, and
  Manning}]{HotpotQA}
Zhilin Yang, Peng Qi, Saizheng Zhang, Yoshua Bengio, William~W. Cohen, Ruslan
  Salakhutdinov, and Christopher~D. Manning. 2018.
\newblock {HotpotQA}: A dataset for diverse, explainable multi-hop question
  answering.
\newblock In \emph{Conference on Empirical Methods in Natural Language
  Processing ({EMNLP})}.

\bibitem[{{Yi} et~al.(2015){Yi}, {Wen-tau}, and Meek}]{WikiQA}
Yang {Yi}, Yih {Wen-tau}, and {Christopher} Meek. 2015.
\newblock \href {https://doi.org/10.18653/v1/D15-1237} {{WikiQA: A Challenge
  Dataset for Open-Domain Question Answering}}.
\newblock page 2013–2018.

\bibitem[{Zellers et~al.(2018)Zellers, Bisk, Schwartz, and Choi}]{swag}
Rowan Zellers, Yonatan Bisk, Roy Schwartz, and Yejin Choi. 2018.
\newblock Swag: A large-scale adversarial dataset for grounded commonsense
  inference.
\newblock \emph{arXiv preprint arXiv:1808.05326}.

\bibitem[{Zellers et~al.(2019)Zellers, Holtzman, Bisk, Farhadi, and
  Choi}]{hellaswag}
Rowan Zellers, Ari Holtzman, Yonatan Bisk, Ali Farhadi, and Yejin Choi. 2019.
\newblock Hellaswag: Can a machine really finish your sentence?
\newblock \emph{arXiv preprint arXiv:1905.07830}.

\bibitem[{Zhang and Tetreault(2019)}]{aelsc}
Rui Zhang and Joel Tetreault. 2019.
\newblock This email could save your life: Introducing the task of email
  subject line generation.
\newblock In \emph{Proceedings of The 57th Annual Meeting of the Association
  for Computational Linguistics}, Florence, Italy.

\bibitem[{Zhang et~al.(2018)Zhang, Liu, Liu, Gao, Duh, and Van~Durme}]{RECORD}
Sheng Zhang, Xiaodong Liu, Jingjing Liu, Jianfeng Gao, Kevin Duh, and Benjamin
  Van~Durme. 2018.
\newblock Record: Bridging the gap between human and machine commonsense
  reading comprehension.
\newblock \emph{arXiv preprint arXiv:1810.12885}.

\bibitem[{Zhang et~al.(2015{\natexlab{a}})Zhang, Zhao, and
  LeCun}]{yelp_polarity}
Xiang Zhang, Junbo Zhao, and Yann LeCun. 2015{\natexlab{a}}.
\newblock \href {http://arxiv.org/abs/1509.01626} {Character-level
  {{Convolutional Networks}} for {{Text Classification}}}.
\newblock \emph{arXiv:1509.01626 [cs]}.

\bibitem[{Zhang et~al.(2015{\natexlab{b}})Zhang, Zhao, and
  LeCun}]{Zhang2015CharacterlevelCN}
Xiang Zhang, Junbo~Jake Zhao, and Yann LeCun. 2015{\natexlab{b}}.
\newblock Character-level convolutional networks for text classification.
\newblock In \emph{NIPS}.

\end{thebibliography}
\bibliographystyle{acl_natbib}

\appendix

\section{Appendices}
\subsection{Datasets Used}
\begin{enumerate}
    \item CoLA \cite{cola}
    \item SST-2 \cite{sst2}
    \item MRPC \cite{mrpc}
    \item QQP \cite{qqp}
    \item MNLI \cite{mnli}
    \item QNLI \cite{qnli}
    \item RTE \cite{rte}
    \item WNLI \cite{winograd}
    \item SuperGLUE \cite{wang2019superglue}
    \item Bool Q \cite{clark2019boolq}
    \item MultiRC \cite{khashabi2018looking}
    \item WIC \cite{pilehvar2018wic}
    \item WSC \cite{levesque2011winograd}
    \item CB \cite{demarneffe:cb}
    \item COPA \cite {roemmele2011choice}
    \item AG News \cite{Zhang2015CharacterlevelCN}
    \item IMDB \cite{IMDB}
    \item MultiNLI \cite{MultiNLI}
    \item SNLI \cite{snli:emnlp2015}
    \item HANS \cite{HANS}
    \item Rotten Tomatoes \cite{RottenTomatoes}
    \item Yelp Polarity \cite{yelp_polarity}
    \item Eraser Multi RC \cite{eraser}
    \item Wiki QA \cite{WikiQA}
    \item Trec  \cite{li-roth-2002-learning, hovy-etal-2001-toward}
    \item SciTail \cite{scitail}
    \item CNN Daily Mail \cite{cnndailymail}
    \item Billsum \cite{billsum}
    \item XSUM \cite{xsum}
    \item Aeslc \cite{aelsc}
    \item Multinews \cite{multinews}
    \item Math QA \cite{mathqa}
    \item Openbook QA \citep{openbookqa}
    \item SWAG \cite{swag}
    \item HellaSWAG \cite{hellaswag}
    \item RACE \cite{race}
    \item CommonSense QA \cite{commonsenseqa}
    \item Cosmos QA \cite{cosmosqa}
    \item AI2 ARC - Easy \cite{arc}
    \item AI2 ARC - Challenge \cite{arc}
    \item SCIQ \cite{sciq}
    \item SQUAD \cite{SQUAD}
    \item NQ \cite{NaturalQuestions}
    \item DROP \cite{DROP}
    \item RECORD \cite{RECORD}
    \item Hotpot \cite{HotpotQA}
    \item TriviaQA \cite{TriviaQA}
    
\end{enumerate}
\label{sec:appendix_dataset}

\subsection{Hyperparameters}
\label{sec:appendix_hp}

\end{document}